\documentclass[10pt,letterpaper]{article}
\usepackage{marvosym}
\usepackage{flushend}
\usepackage{cogsci}
\usepackage{pslatex}
\usepackage{apacite}
\usepackage{natbib}
\usepackage{graphicx}
\usepackage{subcaption}
\PassOptionsToPackage{hyphens}{url}
\usepackage[draft]{hyperref}

\newcommand\blfootnote[1]{%
  \begingroup
  \renewcommand\thefootnote{}\footnote{#1}%
  \addtocounter{footnote}{-1}%
  \endgroup
}

\title{Same-different problems strain convolutional neural networks}
 
\author{{\large \bf Matthew Ricci \textsuperscript{$\dagger$, *} (\texttt{matthew$\_$ricci$\_$1 at brown.edu})}\\ 
  \AND {\large \bf Junkyung Kim \textsuperscript{$\dagger$, *} (\texttt{junkyung$\_$kim at brown.edu})}\\ 
  \AND  {\large \bf Thomas Serre \textsuperscript{*}(\texttt{thomas$\_$serre at brown.edu})}}

\begin{document}

\maketitle

\begin{abstract}
The robust and efficient recognition of visual relations in images is a hallmark of biological vision. We argue that, despite recent progress in visual recognition, modern machine vision algorithms are severely limited in their ability to learn visual relations. Through controlled experiments, we demonstrate that visual-relation problems strain convolutional neural networks (CNNs). The networks eventually break altogether when rote memorization becomes impossible, as when intra-class variability exceeds network capacity. Motivated by the comparable success of biological vision, we argue that feedback mechanisms including attention and perceptual grouping may be the key computational components underlying abstract visual reasoning.

\end{abstract}\vspace{-2mm}

\blfootnote{To appear in the Proceedings of the Annual Meeting of the Cognitive Science Society, 2018} \blfootnote{\textsuperscript{$\dagger$} These authors contributed equally to this work.} \blfootnote{$^*$ Brown University, Department of Cognitive, Linguistic and Psychological Sciences. 190 Thayer Street. Providence, RI 029012} 
\enlargethispage{3mm}

\vspace{-5mm}\section{Introduction}
\begin{figure}[t!]
    \centering
    \includegraphics[width=.45\textwidth]{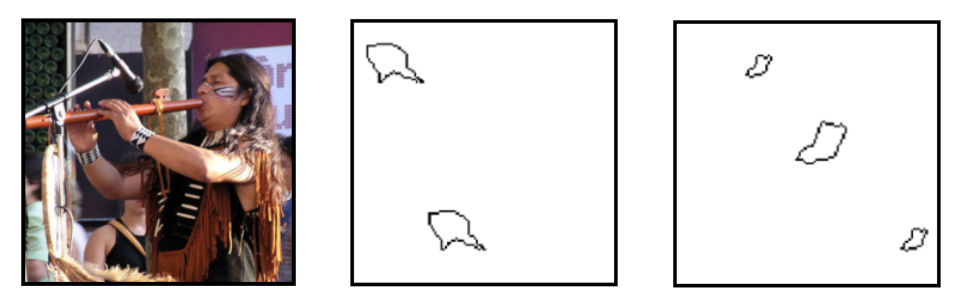}\vspace{-2mm}
    \caption{\emph{Three images.} The image in the left panel can be classified confidently as containing a flute by modern vision algorithms. However, these same algorithms struggle to learn the concept of ``sameness'' as exemplified by the image with the two curves shown in the middle panel. The image in the right panel panel depicts a spatial relation: three objects arranged in a line with the largest in the middle. Middle and are right images are from SVRT  \citep{Fleuret2011}.} 
\label{fig:flute}\vspace{-3mm}
\end{figure}

\noindent Consider the images in Fig.~\ref{fig:flute}. The image on the left was correctly classified as a flute by a deep convolutional neural network~\citep[CNN;][]{He2015a}. This is quite a remarkable feat for such a complicated image. After the network was trained on millions of photographs, this and many other images were accurately categorized into one thousand natural object categories, surpassing, for the first time, the accuracy of a human observer on the ImageNet classification challenge. 

Now, consider the image in the middle. On its face, it is quite simple compared to the image on the left. It is just a binary image containing two curves. Further, it has a rather distinguishing property, at least to the human eye: both curves are the same. The relation between the two items in this simple scene is rather intuitive and immediately obvious to a human observer. Yet, the CNN failed to learn this relation even after seeing millions of training examples. 

Why is it that a CNN can accurately detect the flute while struggling to recognize the simple relation depicted in the middle panel of Fig.~\ref{fig:flute}? That such task is extremely difficult for contemporary computer vision algorithms like CNNs, is known \citep{Fleuret2011, Gulcehre2013, Ellis2015, Stabinger2016}. However, these results, which often relied on a single architecture, were not entirely conclusive: does the inability of CNNs to solve various visual-relation problems reflect a poor choice of network hyperparameters or rather a systematic failure of the entire class of models? To our knowledge, there has been no systematic exploration of the limits of contemporary machine learning algorithms on relational reasoning problems.
\newline
\indent In this study, we will probe the limits of CNNs on visual-relation tasks. In Experiment 1, we perform a systematic performance analysis of CNN architectures on each of the twenty-three synthetic visual reasoning test (SVRT) problems, which reveals a dichotomy of visual-relation tasks: hard same-different problems vs. easy spatial-relation problems. In Experiment 2, we describe a novel, controlled, visual-relation challenge which convincingly shows that CNNs solve same-different tasks via rote memorization. With these experiments, we hope to motivate the computer vision community to reconsider existing visual question answering challenges and turn to cognitive science and neuroscience for inspiration in the design of visual reasoning architectures. 
\enlargethispage{3mm}

\section{Experiment 1: SVRT}
The synthetic visual reasoning test (SVRT) is a collection of twenty-three binary classification problems in which opposing classes differ based on whether their stimuli obey an abstract rule \citep{Fleuret2011}. For example, in problem number 1, positive examples feature two items which are the same up to translation (Fig.~\ref{fig:flute}, middle panel), whereas negative examples do not. In problem 9, positive examples have three items, the largest of which is in between the two smaller ones (Fig.~\ref{fig:flute}, right panel). All stimuli depict simple, closed, black curves on a white background. 


\vspace{1mm}\noindent \textbf{\emph{Methods}}. We tested nine different CNNs of three different depths (2, 4 and 6 convolutional layers) and with three different convolutional filter sizes (2$\times$2, 4$\times$4 and 6$\times$6) in the first layer. This initial receptive field size effectively determines the size of receptive fields throughout the network. The number of filters in the first layer was 6, 12 or 18, respectively, for each choice of initial receptive field size. In the other convolutional layers, filter size was fixed at 2$\times$2 with the number of filters doubling every layer. All convolutional layers had strides of 1 and used ReLU activations. Pooling layers were placed after every convolutional layer, with pooling kernels of size 3$\times$3 and strides of 2. On top of the retinotopic layers, all nine CNNs had three fully connected layers with 1,024 hidden units in each layer, followed by a 2-dimensional classification layer. All CNNs were trained on all problems. Network parameters were initialized using Xavier initialization \citep{glorot10a} and were trained using the Adaptive Moment Estimation (Adam) optimizer \citep{Kingma2015} with base learning rate of $\eta = 10^{-4}$. All experiments were run using TensorFlow \citep{Abadi2016}.
\newline
\begin{figure}[t]
    \centering
    \includegraphics[width=.5\textwidth]{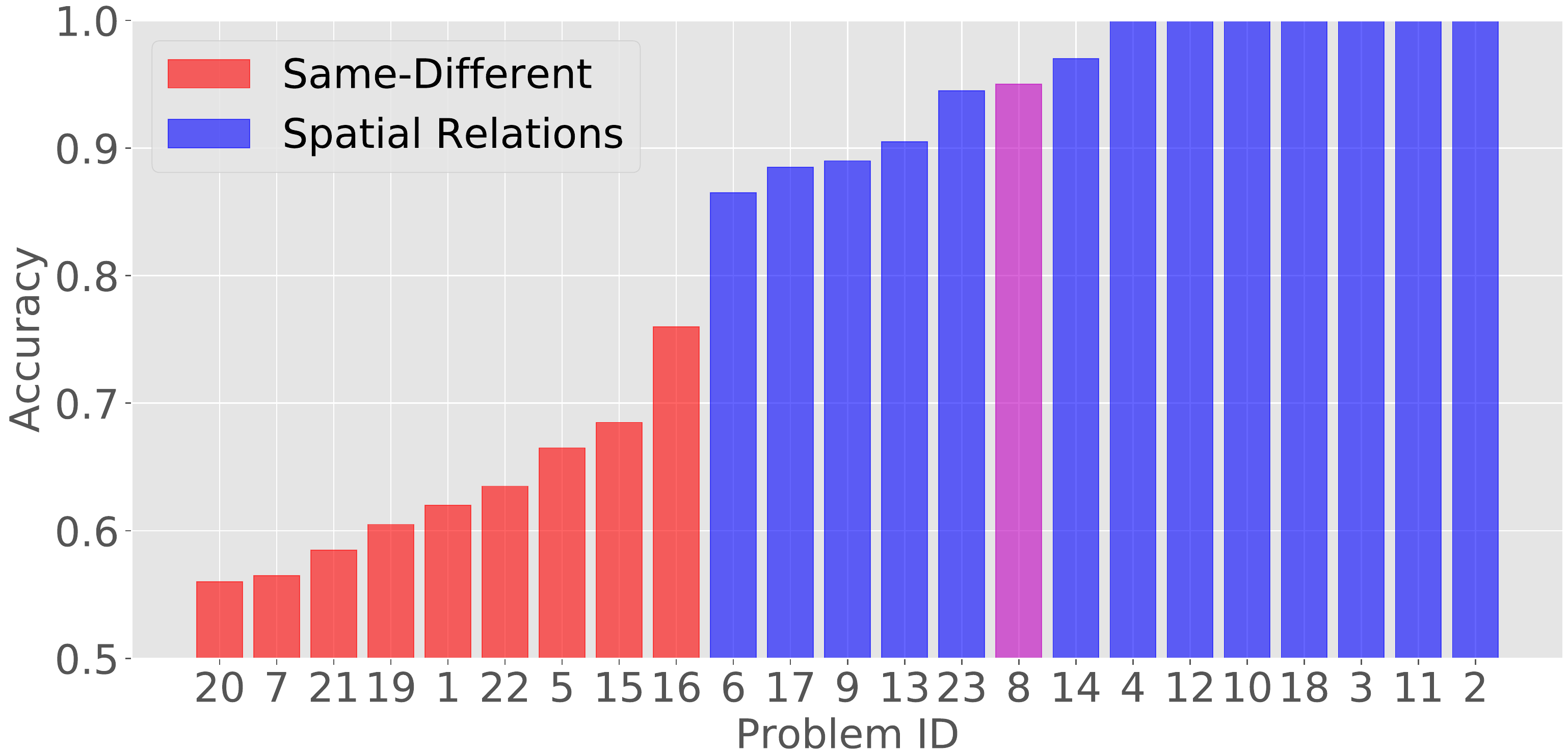}
    \vspace{-5mm}
    \caption{\emph{SVRT results}. Multiple CNNs were trained on each of the twenty-three SVRT problems. Shown are the ranked accuracies of the best-performing network for each problem. The $x$-axis shows the problem ID. CNNs were found to produce uniformly lower accuracies on same-different problems (red bars) than on spatial-relation problems (blue bars). The purple bar represents a problem which required detecting both a same-different relation and a spatial relation.}
    \label{svrt_bars}
\end{figure}
\enlargethispage{3mm}
\indent We obtained the accuracy from the best network for each problem individually. Then, we organized the results into a bar plot, sorted the problems by accuracy and colored the bars red or blue according to the SVRT problem descriptions in~\citep{Fleuret2011}. Problems whose descriptions had words like ``same" or ``identical'' were colored red. These \emph{Same-Different} (SD) problems had items that are congruent up to some transformation (e.g., middle panel, Fig.~\ref{fig:flute}). \emph{Spatial-Relation} (SR) problems, whose descriptions have phrases like ``left of'', ``next to'' or ``touching,'' were colored blue (e.g., right panel, Fig.~\ref{fig:flute}). 

\vspace{1mm} \noindent \textbf{\emph{Results}}. The resulting dichotomy across the SVRT problems is striking (Fig.~\ref{svrt_bars}). CNNs fare uniformly worse on SD problems than they do on SR problems. Many SR problems were learned satisfactorily, whereas some SD problems (e.g., problems 20 and 7) resulted in accuracy not substantially above chance. From this analysis, it appears as if SD tasks pose a particularly difficult challenge to CNNs. This result matches earlier evidence for a visual-relation dichotomy hypothesized by \citet{Stabinger2016}. Additionally, our search revealed that SR problems are equally well-learned across all network configurations, with less than 10\% difference in final accuracy between the worst case and the best case. On the other hand, larger networks yielded significantly higher accuracy than smaller ones on SD problems, suggesting that SD problems are more capacity-sensitive than SR problems. Experiment 1 corroborates earlier studies \citep{Fleuret2011, Gulcehre2013, Ellis2015, Santoro2017} which found that CNNs perform badly on many visual-relation problems and additionally suggests that low performance cannot be simply attributed to a poor choice of hyperparameters. 

\enlargethispage{3mm}

\begin{figure}[t!]
\begin{minipage}[r]{0.45\textwidth}
    \centering
   \includegraphics[scale=.6]{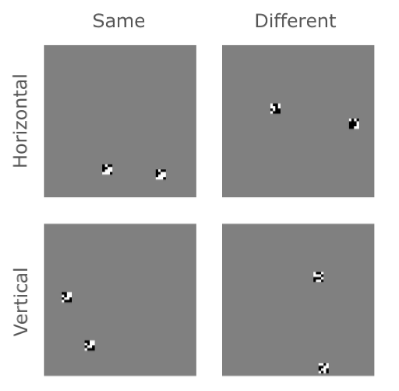}
\caption{\emph{The PSVRT challenge}. Four images show the joint categories of SD and SR problems. An image is \emph{Same} or \emph{Different} depending on whether it contains identical (left column) or different (right column) square bit patterns. An image is \emph{Horizontal} or \emph{Vertical} depending on the average angular displacement between the items.}
\label{fig:psvrt_categories}
\end{minipage}
\end{figure}
\vspace{-5mm}
\section{Experiment 2: PSVRT}
Though useful for surveying many types of relations, the SVRT challenge has two important limitations. First, different problems have different visual structure; e.g., problem 1 requires that an image have two items (Fig. ~\ref{fig:flute}, middle), while problem 9 requires that an image have three (Fig. ~\ref{fig:flute}, right). Therefore, image features, not abstract relational rules, might make some problems harder than others. Second, the ad hoc procedure used to generate simple, closed curves as items in SVRT prevents quantification of image variability and its effect on task difficulty. As a result, even within a single problem in SVRT, it is unclear whether its difficulty is inherent to the classification rule itself or rather the choice of image generation parameters unrelated to the rule. 

To address these limitations, we constructed a new visual-relation benchmark consisting of two idealized problems (Fig.~\ref{fig:psvrt_categories}) from the dichotomy that emerged from Experiment 1: \emph{Spatial Relations} (SR) and \emph{Same-Different} (SD). Critically, both problems used exactly the same images, but with different labels. Further, we parameterized the dataset so that we could systematically control the size of scene items, the number of scene items, and the size of the whole image. Items were binary bit patterns placed on a blank background. 

For each configuration of image parameters, we trained a new instance of a single CNN architecture and measured the ease with which it fit the data. Our goal was to examine how hard it is for a CNN architecture to learn relations for visually different but conceptually equivalent problems. If CNNs can truly learn the ``rule'' underlying these problems, then one would expect the models to learn all problems with more-or-less equal ease. However, if the CNNs only memorize the distinguishing features of the two image classes, then learning should be affected by the variability of the example images in each category. For example, when image size and items size are large, there are simply more possible samples, which might put a strain on the representational capacity of a CNN trying to learn by rote memorization. 

\enlargethispage{3mm}
\noindent \textbf{\emph{Methods}}. Our image generator uses three parameters to control image variability: the size ($m$) of each bit pattern or item, the size ($n$) of the input image and the number ($k$) of items in an image. Our parametric construction allows a dissociation between two possible factors that may affect problem difficulty: classification rules vs. image variability. To highlight the parametric nature of the images, we call this new challenge the \emph{parametric SVRT} or \emph{PSVRT}.
\newline
\indent The image generator is designed such that each image can be used to pose both problems by simply labeling it according to different rules (Fig.~\ref{fig:psvrt_categories}). In SR, an image is classified according to whether scene items are arranged horizontally or vertically as measured by the orientation of the line joining their centers (with a $45^{\circ}$ threshold). In SD, an image is classified according to whether or not it contains at least two identical items. When $k \geq 3$, the SR category label is determined according to whether the average orientation of the displacements between all pairs of items is greater than or equal to $45^{\circ}$. Each image can be labeled according to either the SR or SD rules, so we can ensure the image distribution is identical between the two problem types.
\newline
\indent We trained the same CNN repeatedly from scratch over multiple subsets of the data in order to see if learnability depends on the dataset's image parameters. Training accuracy was sampled at regular intervals and samples were averaged across the length of a training run as well as over multiple trials for each condition, yielding a scalar measure of learnability called ``mean area under the learning curve'' (mean ALC). ALC is high when accuracy increases earlier and more rapidly throughout the course of training and/or when it converges to a higher final accuracy by the end of training. 
\newline
\indent First, we found a baseline architecture which could easily learn both same-different and spatial-relation PSVRT problems for one parameter configuration (item size $m = 4$, image size $n = 60$ and item number $k=2$). Then, for a range of combinations of item size, image size and number of items, we trained an instance of this architecture from scratch.
\newline 
\indent The baseline CNN we used in this experiment had four convolutional layers. The first layer had 8 filters with a 4$\times$4 receptive field size. In the rest of convolutional layers, filter size was fixed at 2$\times$2 with the number of filters in each layer doubling from the immediately preceding layer. All convolutional layers had ReLU activations with strides of 1. Pooling layers were placed after every convolutional layer, with pooling kernels of size 3$\times$3 and strides of 2. On top of retinotopic layers were three fully connected layers with 256 hidden units each, followed by a 2-dimensional classification layer. We initialized all parameters with the Xavier method, optimized the network with Adam with base rate $\eta=10^{-4}$ and ran all experiments in Tensorflow.
\newline
\indent To understand the effect of network size on learnability, we also used two control networks in this experiment: (1) a ``wide" control that had the same depth as the baseline but twice as many filters in the convolutional layers and four times as many hidden units in the fully connected layers and (2) and a ``deep" control which had twice as many convolutional layers as the baseline, by adding a convolutional layer of filter size 2$\times$2 after each existing convolutional layer. Each extra convolutional layer had the same number of filters as the immediately preceding convolutional layer.
\newline
\indent We separately varied the three image parameters to examine their effects on learnability. This resulted in three sub-experiments ($n$ was varied between 30 and 180 while $m$ and $k$ were fixed at 4 and 2, respectively; $m$ was varied between 3 and 7, while $n$ and $k$ were fixed at $60$ and $2$, respectively; $k$ was varied between 2 and 6 while $n$ and $m$ were fixed at 60 and 4, respectively). The baseline CNN was trained from scratch in each condition with 20 million training images and a batch size of 50. 

\enlargethispage{3mm}

\begin{figure*}[t!]
\centering
    \includegraphics[width=1\textwidth]{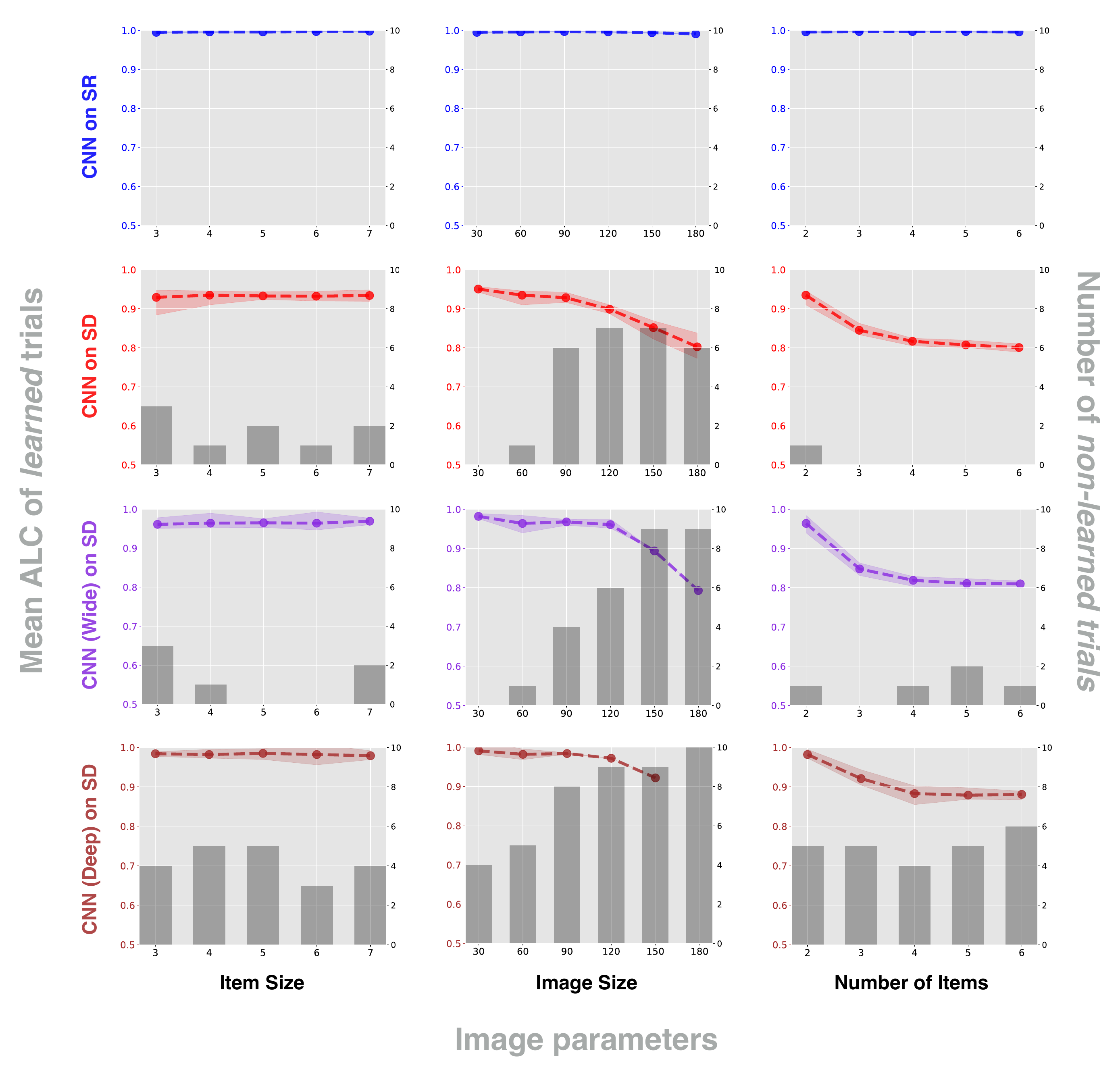}\vspace{-.5cm}
	\caption{\emph{Mean area under the learning curve (ALC) over PSVRT image parameters.} ALC is the normalized area under a training accuracy curve over 20 million images. Colored dots are the mean ALCs for learned trials (in which validation accuracy exceeded 55\%) out of 10 randomly initialized trials. Shaded regions around the colored dots indicate the intervals between the maximum and the minimum ALC among learned trials. Gray bars denote the number of non-learned trials, out of 10 trials. Three model-task combinations (CNN on SR (blue), CNN on SD (red), wide CNN control on SD (violet) and deep CNN control on SD (brown)) are plotted, and each combination is shown for three image parameters: item size, image size and number of items.}
	\label{fig:psvrt_results}
\end{figure*}

\noindent \textbf{\emph{Results}}. In all cases where learning occurred, training accuracy eventually jumped from chance-level and gradually plateaued. In other cases, accuracy remained at chance throughout a training session and the ALC was 0.5. Within a single condition, the CNN often only learned for a fraction of 10 randomly initialized trials. This led us to use two different quantities for describing a model's performance: (1) mean ALC obtained from \emph{learned} trials (in which accuracy crossed $55\%$) and (2) the number of trials in which the learning event never took place (\emph{non-learned}). Note that these two quantities are independent, computed from two complementary subsets of 10 trials.
\newline
\indent In all conditions, we found a strong dichotomy between SD and SR conditions. In SR, across all image parameters and in all trials, the model immediately learned at the start of training and quickly approached 100\% accuracy, producing consistently high and flat mean ALC curves (Fig.~\ref{fig:psvrt_results}, blue dotted lines). In SD, however, we found that the overall ALC was significantly lower than SR (Fig.~\ref{fig:psvrt_results}, red dotted lines). 
We also identified two ways in which image variability affects learnability. First, among the trials in which learning occurred, the final accuracy achieved by the CNN decreased as image size ($n$) and number of items ($k$) increased. This caused ALC to decrease from around 0.95 to 0.8. Second, increasing $n$ also decreased the chance of learning altogether, with more than half of the trials failing to escape chance level when image size was greater than 60 (Fig.~\ref{fig:psvrt_results}, gray bars). In contrast, increasing item size never strained CNN performance. Similar to SR, learnability, both in terms of the proportion of successful trials as well as final accuracy, did not change significantly over the range of item sizes. 
\newline
\indent The fact that straining is only observed in SD, and not in SR and that it is only observed along some of the image parameters, $n$ and $k$, suggests that  straining is not simply a direct outcome of an increase in image variability. Using a CNN with more than twice the number of kernels (Fig.~\ref{fig:psvrt_results}, purple dotted lines) or with twice as many convolutional layers (Fig.~\ref{fig:psvrt_results}, brown dotted lines) as the control did not qualitatively change the trend observed in the baseline model. Although increasing network size did result in improved learned accuracy in general, it also made learning less likely, yielding more non-learned trials than the baseline CNN.
\newpage
\indent We also rule out the possibility of the loss of spatial acuity from pooling or subsampling operations as a possible cause of straining. Our CNNs achieved the best overall accuracy when image size was smallest. If the loss of spatial acuity was the source of straining, increasing image size should have improved the network's performance instead of hurting it because items would have tended to be placed farther apart from each other. Moreover, in other experiments \citep{Kim2018}, we found that networks with identical spatial acuity exhibited no straining as long as items were segregated into different channels.
\newline
\indent The weak effects of item size and item number shed light on the computational strategy used by CNNs to solve SD. We hypothesize that CNNs learn ``subtraction templates'', filters with one positive region and one negative region (like a Haar or Gabor wavelet), in order to detect the similarity between two image regions. A different subtraction template is required for each relative arrangement of items, since each item must lie in one of the template's two regions. When identical items lie in these opposing regions, they are subtracted by the synaptic weights. This difference is then used to choose the appropriate same/different label. This strategy does not require memorizing specific items, so increasing item size (and therefore total number of possible items) should not make the task appreciably harder. Further, a single subtraction template can be used even in scenes with more than two items, since images are classified as ``same'' when they have \emph{at least} two identical items. So, any straining effect from item number should be negligible as well. Instead, the principal straining effect with this strategy should arise from image size, which exponentially increases the possible number arrangements of items. 

Taken together, these results suggest that, when CNNs learn a PSVRT problem, they are simply building a feature set tailored to the relative positional arrangements of items in a particular data set, instead of learning the abstract ``rule'' per se. 
\enlargethispage{3mm}
\vspace{-4mm}
\section{Discussion}
Our results indicate that visual-relation problems can quickly exceed the representational capacity of feedforward networks. While learning templates for individual objects appears to be tractable for today's deep networks, learning templates for \emph{arrangements} of objects becomes rapidly intractable because of the combinatorial explosion in the requisite number of features to be stored. That stimuli with a combinatorial structure are difficult to represent with feedforward networks has been long acknowledged by cognitive scientists \citep{Fodor1988}.
\newline
\indent Compared to the feedforward networks in this study, biological visual systems excel at detecting relations. \citet{Fleuret2011} found that humans can learn rather complicated visual rules and generalize them to new instances from just a few SVRT training examples. Their participants could learn the rule underlying the hardest SVRT problem for CNNs in our Experiment 1, problem 20, from an average of about 6 examples. Problem 20 is rather complicated, involving two shapes such that ``one shape can be obtained from the other by reflection around the perpendicular bisector of the line joining their centers." In contrast, the best performing network for this problem could not get significantly above chance after one million training examples.  
\newline
\enlargethispage{3mm}
\indent Visual reasoning ability is not just found in humans. Birds and primates can be trained to recognize same-different relations and then transfer this knowledge to novel objects \citep{Wright2006}. A striking example of same-different learning in animals comes from \citet{MartinhoIII2016} who showed that newborn ducklings can learn the abstract concept of sameness from a single example. In contrast, we have found in follow-up work that state-of-the-art neural networks demonstrated no ability to transfer the concept of same-different to novel objects even after hundreds of thousands of training examples \citep{Kim2018}.
\newline
\indent It is relatively well accepted that, despite the widespread presence of feedback connections in our visual cortex, certain visual recognition tasks, including the detection of natural object categories, are possible in the near absence of cortical feedback -- based primarily on a single feedforward sweep of activity through our visual cortex \citep{Serre2016-vm}. However, psychophysical evidence suggests that this feedforward sweep is too spatially coarse to localize objects even when they can be recognized \citep{Evans2005}. The implication is that object localization in clutter requires attention \citep{Zhang2011-mo}. It is difficult to imagine how one could recognize a relation between two objects without spatial information. Indeed, converging evidence  \citep{Logan1994,Moore1994,Rosielle2002,Holcombe2011,Franconeri2012,vanderham2012} suggests that the processing of spatial relations between pairs of objects in a cluttered scene requires attention, even when individual items can be detected pre-attentively.
\newline
\indent In follow-up work \citep{Kim2018}, we argued that perceptual grouping, a mechanism for binding features into discrete objects \citep{Roelfsema2006}, is another key non-feedforward process supporting visual relation detection. We found that relational networks \citep{Santoro2017}, CNN extensions that exhaustively attend to all unbound features in a deep layer, are strained just like CNNs and tend to easily overfit. In contrast, we showed that a network which simulates the effects of perceptual grouping by forcing scene items into separate channels can easily learn our PSVRT tasks without straining. This toy network simulates in a feedforward manner the dynamic sequence of attention shifts between perceptually grouped features believed to underlie visual relation detection \citep{Franconeri2012}. These dynamic representations built ``on-the-fly" circumvent the combinatorial explosion associated with the storage of synaptic templates for all possible relations, helping to prevent the capacity overload associated with feedforward neural networks. 
\newline
\indent Humans can easily detect when two objects are the same up to some transformation \citep{Shepard1971} or when objects exist in a given spatial relation \citep{Fleuret2011, Franconeri2012}. More generally, humans can effortlessly construct an unbounded set of structured descriptions about their visual world \citep{Geman2015}. Given the vast superiority of humans over modern computers in their ability to detect visual relations, we see the exploration of attentional and grouping mechanisms as an important next step in our computational understanding of visual reasoning. 
\vspace{-3mm}
\section{Acknowledgments}
The authors would like to thank Drs. Drew Linsley and Sven Eberhardt for their advice, along with Dan Shiebler for earlier work. This research was supported by NSF early career award (IIS-1252951) and DARPA young faculty award (YFA N66001-14-1-4037). Additional support was provided by the Center for Computation and Visualization (CCV) at Brown University. This material is based upon work supported by author MR's National Science Foundation Graduate Research Fellowship under Grant No. 1644760.
\vspace{-2.5mm}
\bibliography{RS_bib}
\bibliographystyle{apacite}

\setlength{\bibleftmargin}{.125in}
\setlength{\bibindent}{-\bibleftmargin}

\end{document}